\documentclass[conference, 9pt]{IEEEtran}
\IEEEoverridecommandlockouts
\usepackage{multirow}     % For \multirow
\usepackage[table,dvipsnames]{xcolor}
\definecolor{Gainsboro}{RGB}{220,220,220}
\usepackage{cite}
\usepackage{caption}
\usepackage{subcaption}
\usepackage{url}
\usepackage{amsmath,amssymb,amsfonts}
\usepackage{algorithmic}
\usepackage{graphicx}
\usepackage{textcomp}
\usepackage{booktabs}
\usepackage{tikz}
\usepackage{float}
\usepackage{placeins}
\def\BibTeX{{\rm B\kern-.05em{\sc i\kern-.025em b}\kern-.08em
    T\kern-.1667em\lower.7ex\hbox{E}\kern-.125emX}}
\begin{document}

\title{WST: Weakly Supervised Transducer for Automatic Speech Recognition}
%{\footnotesize \textsuperscript{*}Note: Sub-titles are not captured for https://ieeexplore.ieee.org  and
%should not be used}
%\thanks{Identify applicable funding agency here. If none, delete this.}
%}

\author{
    \IEEEauthorblockN{Dongji Gao$^{\star}$, Chenda Liao$^\dagger$, Changliang Liu$^\dagger$, Matthew Wiesner$^{\star}$,}
    \IEEEauthorblockN{Leibny Paola Garcia$^{\star}$, Daniel Povey$^\ddagger$, Sanjeev Khudanpur$^{\star}$, Jian Wu$^\dagger$}
    \IEEEauthorblockA{\textit{Johns Hopkins University$^{\star}$, Xiaomi$^\ddagger$} \\
    \textit{Microsoft$^\dagger$} \\
    \textit{dgao5@jhu.edu}
}
}

\maketitle

\begin{abstract}
The Recurrent Neural Network-Transducer (RNN-T) is widely adopted in end-to-end (E2E) automatic speech recognition (ASR) tasks but depends heavily on large-scale, high-quality annotated data, which are often costly and difficult to obtain. To mitigate this reliance, we propose a Weakly Supervised Transducer (WST), which integrates a flexible training graph designed to robustly handle errors in the transcripts without requiring additional confidence estimation or auxiliary pre-trained models. Empirical evaluations on synthetic and industrial datasets reveal that WST effectively maintains performance even with transcription error rates of up to $70\%$, consistently outperforming existing Connectionist Temporal Classification (CTC)-based weakly supervised approaches, such as Bypass Temporal Classification (BTC) and Omni-Temporal Classification (OTC). These results demonstrate the practical utility and robustness of WST in realistic ASR settings. The implementation will be publicly available.
\end{abstract}


\section{Introduction}
Automatic Speech Recognition (ASR) models have seen significant advances in recent years, largely driven by improved model architectures~\cite{Graves2006ConnectionistTC, chan2016listen, graves2012sequence, watanabe2017hybrid, li2022recent} and increased availability of training data~\cite{Li2020OnTC, Lu2020ExploringTF, Wang2020TransformerIA}.
Among these, the Recurrent Neural Network-Transducer (RNN-T)~\footnote{In this paper, the terms ``RNN-T'' and ``Transducer'' are used interchangeably.} has emerged as a popular model, particularly for end-to-end ASR tasks due to its leading performance and ability to learn a direct mapping from acoustic to text without depending on separate acoustic and language models.

Despite its demonstrated effectiveness, an RNN-T is notably data-hungry, often requiring tens of thousands of hours of carefully annotated audio to reach state-of-the-art performance. 
Two considerations make the \textbf{quantity} and \textbf{quality} of training data especially important. 
First, an RNN-T model typically comprises three core components: an encoder, a prediction network, and a joint network. 
Because the model structure is complicated and the task is inherently complex, collecting large-scale, diverse datasets is essential for robust generalization. 
Second, like most ASR models, the RNN-T formulation is built on the assumption that each transcript perfectly matches the corresponding speech. In other words, the model relies on the training data having accurate labels.

However, in real-world scenarios, this assumption rarely holds. 
There is always a trade-off between data quantity and quality. On the one hand, high-quality data often come from prepared read-speech recordings or are transcribed by human annotators, both of which are time-consuming and expensive processes. 
As a result, the high quality datasets are typically limited, especially for low-resource languages and dialects.
On the other hand, to scale model training, researchers often turn to lower quality data sources~\cite{radford2023robust, pratap2024scaling, peng2023reproducing}. 
The transcripts for these recordings may be loosely aligned closed captions or automatically generated labels, which are prone to inaccuracies. 
Although such data can increase the volume of training data, they pose challenges to the RNN-T model, particularly when the assumption of perfect supervision is no longer valid.

Training with erroneous transcripts can significantly degrade ASR performance~\cite{wctc, stc, gao2023bypass, gao2023learning}. Specifically, for transducer models, using incorrect transcripts results in flawed alignment between the acoustic features and the target text, ultimately leading to poor learning outcomes~\cite{laptev2023powerful, keren2024token}. 

Recent research addresses this challenge by leveraging \textit{Weakly Supervised Training} techniques, which allow ASR models to be trained directly on low-quality data. One line of work focuses on improving Connectionist Temporal Classification (CTC) criterion to automatically detect and tolerate misaligned or erroneous transcript segments, such as Wild-card CTC (W-CTC)~\cite{wctc}, Star Temporal Classification (STC)~\cite{stc}, Bypass Temporal Classification (BTC)~\cite{gao2023bypass}, Alternative Pseudo-Labeling (APL)~\cite{apl}, and Omni-Temporal Classification (OTC)~\cite{gao2023learning}. 
Given that RNN-T has demonstrated advantages over CTC in many benchmarks, recent efforts have shifted toward adapting the RNN-T architecture to noisy transcripts. 
Examples include the W-Transducer~\cite{laptev2023powerful} and the Token-Weighted RNN-T~\cite{keren2024token}. 

Building on this direction, we propose a Weakly Supervised Transducer (WST) that addresses these shortcomings by 
introducing a flexible training graph that explicitly accounts for potential transcript errors in the differentiable Weighted Finite-State Transducers (WFST) framework~\cite{povey2016purely}. 
Through this flexible alignment mechanism, the model naturally learns to downweight or bypass unreliable transcript segments. 
As a result, WST not only handles all forms of transcript noise, but also can be trained from scratch without requiring additional confidence estimators or pre-trained ASR systems, making it a straightforward replacement for a standard RNN-T.

\section{Preliminaries}
\subsection{Transducer}
\label{ssec:transducer}
Given the acoustic feature sequence $\mathbf{x}$ of length $T$ and the output transcript $\mathbf{y}$ of length $U$, the transducers compute the posterior probability $P(\mathbf{y} | \mathbf{x})$ by marginalizing over all possible alignments $\mathbf{a}$ between $\mathbf{x}$  and $\mathbf{y}$:
\begin{equation}
    P(\mathbf{y} | \mathbf{x}) = \sum_{\mathbf{a} \in \mathcal{B}^{-1}(\mathbf{y})} P(\mathbf{a} | \mathbf{x}).
\end{equation}
Here, the function $\mathcal{B}$ “collapses” each alignment $\mathbf{a}$ to $\mathbf{y}$. Therefore, $\mathcal{B}^{-1} (\mathbf{y})$ presents the set of all possible alignments between $\mathbf{x}$ and $\mathbf{y}$. 
This set can be visualized as paths of a $T \times U$ lattice.

\begin{figure}[t]
    \centering
    \includegraphics[width=0.6\linewidth]{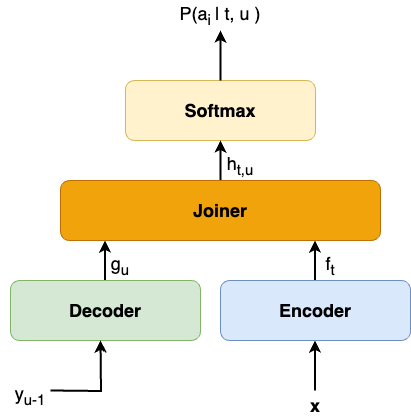}
    \caption{Architecture of RNN-T. It contains an encoder, a decoder, and a joiner followed by a classification layer.}
    \label{fig:chap_1_rnnt_architecture}
\end{figure}

Transducers further parameterize $P(\mathbf{a}|\mathbf{x})$ by introducing an encoder, decoder, and a joiner network, as shown in Figure~\ref{fig:chap_1_rnnt_architecture}. 
Given $\mathbf{x}$, the encoder transforms it into high-level feature representations $\mathbf{f} = [f_{1}, f_{2},\cdots,f_{T}]$. Meanwhile, the decoder processes $\mathbf{y}$ into predictions $\mathbf{g} = [g_{1}, g_{2},\cdots,g_{U}]$. The joiner network then combines these two representations followed by the classification layer:
\begin{align}
    \mathbf{f} &= \text{Encoder} (\mathbf{x}) \\
    \mathbf{g}^{u} &= \text{Decoder} (\mathbf{y}^{u-1}) \\
    h_{t, u} &= \text{Joiner}(f_{t}, g_{u}) \\
    P(a_{i} | \mathbf{x}, \mathbf{y}^{u-1}) &= \text{Softmax} (\mathbf{W} h_{t, u} + b).
\end{align}

Each symbol in an alignment $\mathbf{a}$ is then predicted sequentially, conditioned on the corresponding encoder and decoder outputs:
\begin{align}
    P(\mathbf{a}|\mathbf{x}) &= \prod_{i=1}^{T+U} P(a_{i}|f_{t(i)}, g_{u(i)}) \\
    &= \prod_{i=1}^{T+U} P(a_{i}|h_{t(i),u(i)}),
\end{align}
where $t(i)$ and $u(i)$ denote the encoder and decoder indices associated with the $i$-th symbol in the alignment. 
During training, the transducer model minimizes the loss, defined as:
\begin{equation}
    \mathcal{L}_{\text{RNN-T}} = -\text{log} P(\mathbf{y} | \mathbf{x}) = -\text{log} \sum_{\mathbf{a} \in \mathcal{B}^{-1}(\mathbf{y})} P(\mathbf{a} | \mathbf{x})
\end{equation}

\subsection{Transducer in WFST framework}
As discussed in Section~\ref{ssec:transducer}, given the acoustic feature sequence $\mathbf{x}$ of length $T$ and the output transcript $\mathbf{y}$ of length $U$, there can be $\binom{T+U}{U}$ alignments in the transducer lattice. 
Computing $\mathcal{L}_{\text{RNN-T}}$ by brute-force summation over all alignments is computationally intractable.
To address this, dynamic programming in the form of the forward-backward algorithm is used for efficient computations.
In practice, this algorithm is often implemented manually in CUDA to efficiently compute the forward and backward variables. 
However,this requires a strong understanding of parallel programming and significant effort to manage GPU-specific issues.

\begin{figure}[t]
  \centering
  \includegraphics[width=\linewidth]{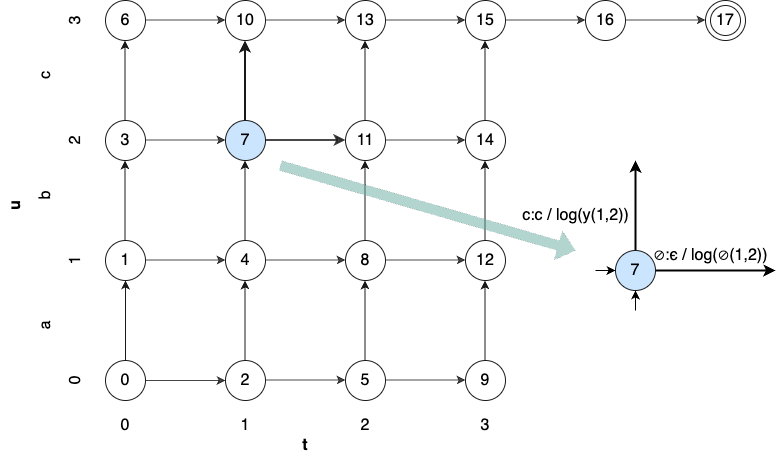}
  \caption{Transducer training graph in the k2 for the transcript ``a b c'' aligned to an input sequence of 4 frames. The graph starts at state 0, and the double-circled state 17 represents the final state. Each arc is labeled with an input symbol and an output symbol (separated by a colon), followed by a weight after the slash indicating the log-probability of emitting the output symbol. State 7 is highlighted as an example. The vertical arc (a \textit{token arc}) emits the output symbol ``c'' without advancing the time step. The horizontal arcs (referred to as \textit{blank arcs}) consumes a time frame but not emitting a label ($\epsilon$).}
  \label{fig:rnnt_wfst}
\end{figure}

Recent research~\cite{laptev2023powerful} has demonstrated that the transducer loss can be easily implemented within the differentiable Weighted Finite State Transducer (WFST) framework, such as k2\footnote{\url{https://github.com/k2-fsa/k2}}.
In k2, each graph state maintains forward and backward variables, which are tracked via the arcs entering and exiting that state. 
Once the WFST graph corresponding to the transducer lattice is constructed, k2 automatically computes the gradients and propagates them back to the model.
The main distinction from a conventional RNN-T lattice is the addition of a final state required by k2, as illustrated in Figure~\ref{fig:rnnt_wfst}.

\section{Method}

We propose an enhancement for WST by introducing additional flexibility to the training graph of a standard transducer model, allowing it to better accommodate transcription uncertainties.

\subsection{Weakly Supervised Transcript Graph}

\begin{figure}[t]
  \centering
  \begin{subfigure}{0.4\linewidth}
    \centering
    \includegraphics[width=0.5cm]{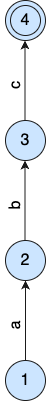}
    \subcaption{Graph of transcript ``a b c".}
    \label{fig:g_a}
  \end{subfigure}
  \begin{subfigure}{0.5\linewidth}
    \centering
    \includegraphics[width=0.87cm]{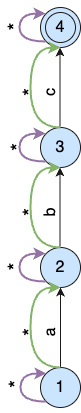}
    \subcaption{Compact graph of ``a b c" and all its variants.}
    \label{fig:g_b}
  \end{subfigure}
  \caption{WFST representation of transcript.}
  \label{fig:g}
\end{figure}

Following the design principles of BTC~\cite{gao2023bypass} and CTC~\cite{gao2023learning}, we employ a WFST to represent the transcript, $\mathbf{y}$, as shown in Figure~\ref{fig:g_a}. 

\begin{figure*}[t]
    \centering
    \includegraphics[width=\linewidth]{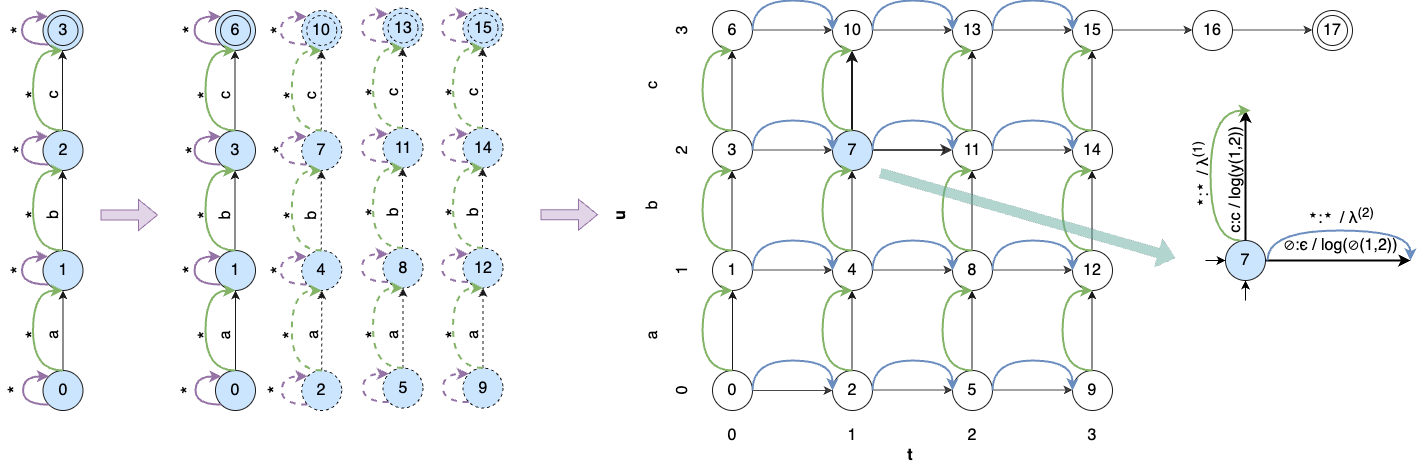}
    \caption{Weakly Supervised Transducer training graph in the k2 for the transcript ``a b c'' aligned to an input sequence of 4 frames. Compared with a standard transducer graph, two types of bypass arcs are added: \textit{token bypass arcs} and \textit{blank bypass arcs}. The token bypass arcs (drawn vertically) enable the model to skip the current token while remaining in the same time frame, whereas the blank bypass arcs (drawn horizontally) allow the insertion of a $\star$ while advancing one time step with certain penalties. For example, at State 7, the \textit{token bypass arc} permits skipping the token ``c'' with penalty $\lambda_{1} $ and the \textit{blank bypass arc} facilitates the insertion of the $\star$ token with penalty $\lambda_{2}$.}
    \label{fig:wsrnnt_wfst}
\end{figure*}

To model all possible transcript errors in $\mathbf{y}$, WST constructs the weakly supervised transcript graph by adopting the compact transcript graph constructions introduced in BTC and OTC. Specifically, we incorporate the special token, $\star$, to abstractly represent model uncertainty (i.e., ``garbage'' tokens). This symbol is incorporated into the WFST through the use of self-loop and bypass arcs to capture a wide range of error patterns—namely, substitution errors, insertion errors, and deletion errors in the transcripts. An example  is shown in Figure~\ref{fig:g_b}.

\subsection{Weakly Supervised Training Graph}

By expanding the compact transcript graph along the time (t) axis, we obtain the weakly supervised transducer training graph, as shown in Figure~\ref{fig:wsrnnt_wfst}.

\begin{figure}[t]
  \centering
  \begin{subfigure}{0.46\linewidth}
    \centering
    \includegraphics[width=2cm]{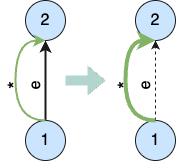}
    \subcaption{Substitution error. The computation of $\mathcal{L}_{\text{RNN-T}}$ can go through $\star$ rather than the incorrect token ``e''.}
    \label{fig:errors_sub}
    \vspace{5mm}
  \end{subfigure}
  \begin{subfigure}{0.46\linewidth}
    \centering
    \includegraphics[width=2cm]{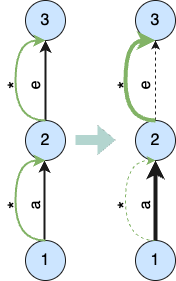}
    \subcaption{Insertion error. The computation of $\mathcal{L}_{\text{RNN-T}}$ can skip over the token ``e'' that was wrongly inserted, while producing the correct neighboring token ``a''.}
    \label{fig:errors_ins}
    \vspace{5mm}
  \end{subfigure}
  \begin{subfigure}{0.4\linewidth}
    \centering
    \includegraphics[width=4cm]{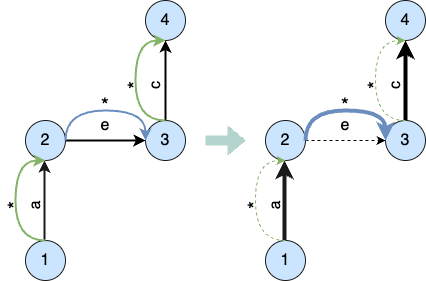}
    \subcaption{Deletion error. The computation of $\mathcal{L}_{\text{RNN-T}}$ can insert the deleted token (between ``a'' and ``c'') by using $\star$ instead of $\oslash$ when transitioning between ``a'' and ``b''.}
    \label{fig:errors_del}
  \end{subfigure}
  \caption{Examples of dealing with different kinds of errors. The thickness of the arc indicates the probability assigned to it.}
  \label{fig:errors}
\end{figure}

Compared to the transducer graph shown in Figure~\ref{fig:rnnt_wfst}, the WST graph extends it by adding two additional types of arcs: \textcolor{green}{token bypass arc} along the transcript (u) axis and \textcolor{blue}{blank bypass arc} along the time (t) axis.

\textit{Token bypass arcs} address substitution and insertion errors. 
For substitution errors, the model can bypass the incorrect token by assigning a higher probability to the token bypass arc, represented by the $\star$ token. 
Similarly, for insertion errors, the model can effectively "absorb" the incorrectly inserted token within the same time frame. 
This mechanism prevents the model from learning incorrect relationships by reducing the impact of erroneous gradient descent during back-propagation\footnote{Unlike in the CTC, BTC, or OTC cases, note that a spurious insertion in the training transcript ${y}$ does not need to be aligned with a minimum number of acoustic frames in $\mathbf{x}$: the Tranducer model can emit the corresponding $\star$ arc without "consuming" an acoustic frame.}. 

\textit{Blank bypass arcs} address deletion errors. In this case, the model can align the orphaned acoustic segments to the token $\star$ on the blank bypass arc instead of $\oslash$, which prevents the model from incorrectly associating acoustics with the blank token. 
This helps avoid emitting too many $\oslash$ tokens during decoding. 
Here, we make the assumption that the number of missing tokens cannot exceed the number of time frames ($T$), which is generally true in ASR scenarios.

Figure~\ref{fig:errors} illustrates the flexible alignment mechanisms enabled by the introduction of the special token $\star$ into the training WFST, specifically through \emph{token bypass arcs} and \emph{blank bypass arcs} designed to address substitution, insertion, and deletion errors.

Figure~\ref{fig:errors_sub} shows a \textbf{substitution error}. Suppose the reference transcript expects token ``a'' at a particular time step, but the corrupted transcript contains token ``e'' instead. The training graph compensates for this by bypassing the mismatched token ``e'' using $\star$.

Figure~\ref{fig:errors_ins} addresses an \textbf{insertion error}, where the transcript contains an extraneous token ``e''. The training WFST allows the model to produce a $\star$ arc from state 2 to 3 (i.e., skip over token ``a'') without aligning it to any frame in $\mathbf{x}$.

Figure~\ref{fig:errors_del} illustrates a \textbf{deletion error}, where a token ``b'' is present in the audio but missing from the transcript. A \textit{blank bypass arc} from state 2 to state 3 allows the model to consume the acoustic corresponding to ``b'' and produce $\star$.

\subsection{Decoder History Approximation}
The decoder in transducer models is typically parameterized in an autoregressive manner, where the hidden states $\mathbf{g}^{u}$ is determined by its previous history $[y_1, \cdots, y_{u-1}]$:
\begin{align}
    \mathbf{g}^{u} &= \text{Decoder} (y_1, \cdots, y_{u-1}).
\end{align}
Under the WST framework, however, incorporating the special token $\star$ into the training graph introduces a complication. At a state $(t, u)$, the transcript history is no longer represented by a linear sequence, but rather a branching graph. For instance, in Figure~\ref{fig:g}, the standard Transducer history at state $2$ would correspond to ``a b'', as shown in Figure~\ref{fig:g_a}. In contrast, the modified graph in Figure~\ref{fig:g_b} allows for multiple alternative histories, such as ``a b'' or ``$\star$ b" or ``a $\star$'' or even ``a $\star$ b''.

To address the issue of diverging histories, we adopt a stateless prediction network (SLP)~\cite{ghodsi2020rnn} for the WST's decoder. 
 In this design, the hidden state $\mathbf{g}_{u}$ depends only on the most recent token:
\begin{align}
    \mathbf{g}_{u} &= \text{Decoder} (y_1, \cdots, y_{u-1}) \\
    &= \text{SLP} (y_{u-1}).
\end{align}
This choice effectively collapses the historical context into only two possibilities for the most recent symbol: either $\star$ or a token from the standard vocabulary $\mathcal{V}$. In the illustration of Figure~\ref{fig:g_b}, at state $3$, the "predictor" context is either ``b'' or $\star$, depending on which arc was used to enter the state.

Moreover, we approximate multiple diverging paths by treating them as if they share the same $\mathbf{g}^{u}$ 
\[
\mathbf{g}^u \;\rightarrow\;
\begin{cases}
    \text{SLP} (\star)\\
    \text{SLP}(y_{u-1})
\end{cases}
= \text{SLP}(y_{u-1}).
\]
This approximation effectively collapses all possible label histories into a single path, enabling the WST decoder to be trained in the same manner as a standard Transducer decoder without introducing additional complexity. This design choice is motivated by the findings of~\cite{ghodsi2020rnn}, which suggest that the encoder and joint networks are sufficient to capture both acoustic and linguistic information, while the prediction network primarily governs the decision to emit a non-blank symbol. Notably,~\cite{ghodsi2020rnn} also shows that when the transcript $\mathbf{y}$ is error-free, training with a stateless prediction network results in minimal or no degradation in ASR performance.

\subsection{Modeling $\star$ token}
The $\star$ token is introduced to represent segments of the transcription that are uncertain or only partially observed. 
Following the OTC strategy, $\star$ is defined as the average probability of all non-blank tokens. Formally,
\begin{align}
 P(\star_{t} \mid \mathbf{x}) = \frac{1 - P(\oslash_{t} \mid \mathbf{x})}{\left| \mathcal{V} \right| - 1} \label{eq:p_wst_star}, 
\end{align}
where $\mathcal{V}$ is the vocabulary.
Therefore, similar to OTC, a forward pass through the neural network with input $\mathbf{x}$ produces frame-level probabilities, which are then summed over to calculate Equation~\ref{eq:p_wst_star}.
\subsection{Arc weight (penalty) strategy}
To reduce hyperparameter tuning overhead, we employ fixed penalties at each training epoch. 
Specifically, for the $i$-th epoch, the penalties $\lambda_{i}^{(1)}$ and $\lambda_{i}^{(2)}$ assigned to arc weights are given by:
\begin{align}
    \lambda_{i}^{(1)} = \beta_{1}\ \quad\text{and}\quad \lambda_{i}^{(2)} = \beta_{2},
\end{align}
where $\beta_{1}$ and $\beta_{2}$ are constant values across epochs, $\lambda^{(1)}$ refers to the penalty for token bypass arcs, and $\lambda^{(2)}$ refers to the penalty for blank bypass arcs (see Figure~\ref{fig:wsrnnt_wfst}).

\section{Experimental Setup}

\subsection{Data Preparation}
We conduct experiments on two English datasets: LibriSpeech~\cite{panayotov2015librispeech} (\texttt{train-clean-100}) and in-house (IH-10k) data.
For the \texttt{train-clean-100} dataset, we employ the same strategy as~\cite{gao2023learning} to generate synthetic errors to data with error rate of $\{0.1, 0.3, 0.5, 0.7\}$. For additional details, please refer to~\cite{gao2023learning}.

IH-10k is an in-house dataset comprising 10,000 hours of industrial training data collected from online sources. Although the data has undergone careful cleaning, transcription errors remain. To quantify the quality of the transcriptions, we manually annotated 400 utterances as ground truth and computed an overall transcription error rate of approximately $10.0\%$, consisting of $4.0\%$ substitution errors, $2.0\%$ insertion errors, and $4.0\%$ deletion errors.

\subsection{Implementation Details}
For feature extraction, we used two different approaches:
\begin{enumerate}
    \item LibriSpeech: We employed the wav2vec 2.0 (base) model~\cite{wav2vec2} to extract 768-dimensional features with a stride of 20 ms.
    \item IH-10k: We extracted 80-dimensional Fbank features to reduce storage.
\end{enumerate}
Our transducer model comprises three main components:
\begin{enumerate}
    \item Encoder: A 12-layer Conformer network~\cite{gulati2020conformer} that converts input audio into high-level acoustic representations.
    \item Decoder: A "stateless" feed-forward network that computes predictions based solely on the current input.
    \item Joiner: A fully connected layer that combines the outputs of the encoder and decoder, followed by a softmax function to generate a probability distribution over the possible tokens.
\end{enumerate}
We use different BPE vocabularies for the two datasets: a vocabulary size of 200 for LibriSpeech and a vocabulary size of 4000 for IH-10k.

\section{Results and Analysis}
\subsection{LibriSpeech}
\begin{figure}[t]
\centering
    \begin{subfigure}[b]{0.49\linewidth}
        \centering
        \includegraphics[width=\linewidth]{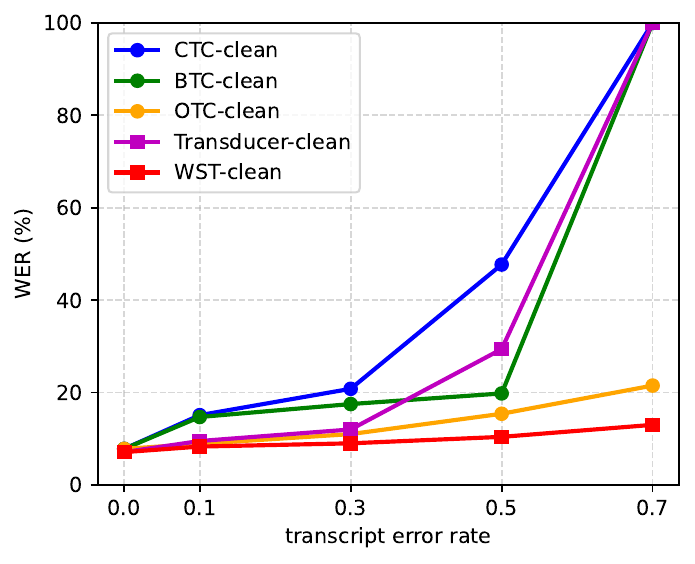}
        \caption{Substitution}
        \vspace{5mm}
    \end{subfigure}
    \begin{subfigure}[b]{0.49\linewidth}
        \centering
        \includegraphics[width=\linewidth]{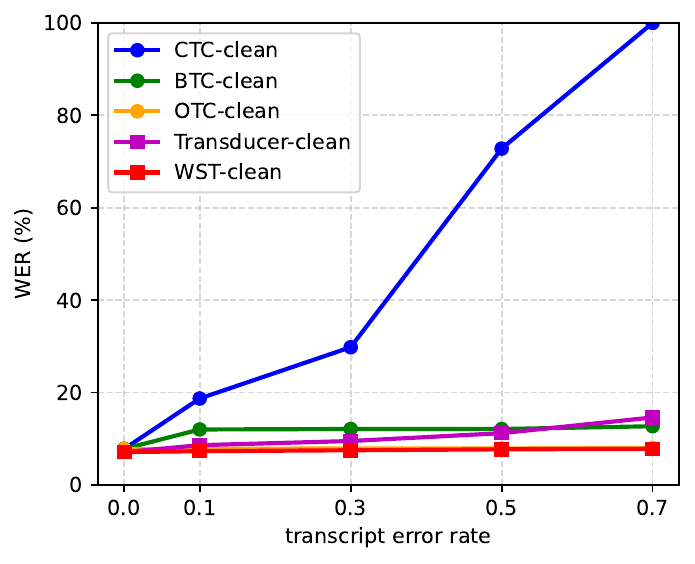}
        \caption{Insertion}
        \vspace{5mm}
    \end{subfigure}\hfill
    \begin{subfigure}[b]{0.49\linewidth}
        \centering
        \includegraphics[width=\linewidth]{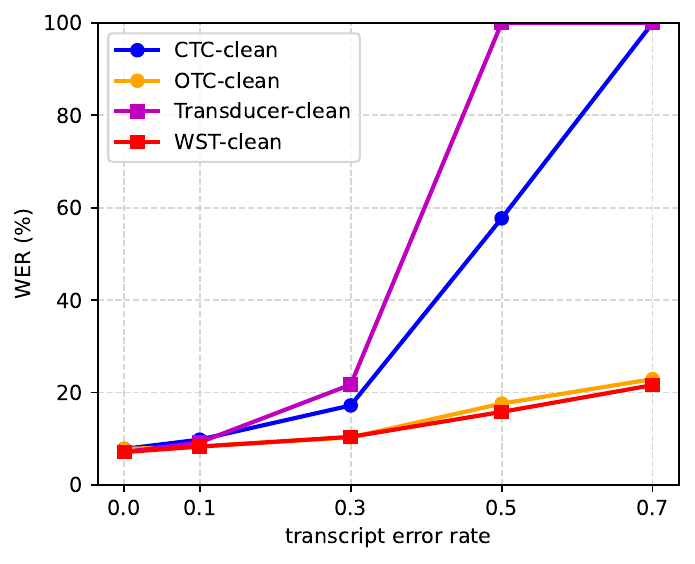}
        \caption{Deletion}
    \end{subfigure}
    \begin{subfigure}[b]{0.49\linewidth}
        \centering
        \includegraphics[width=\linewidth]{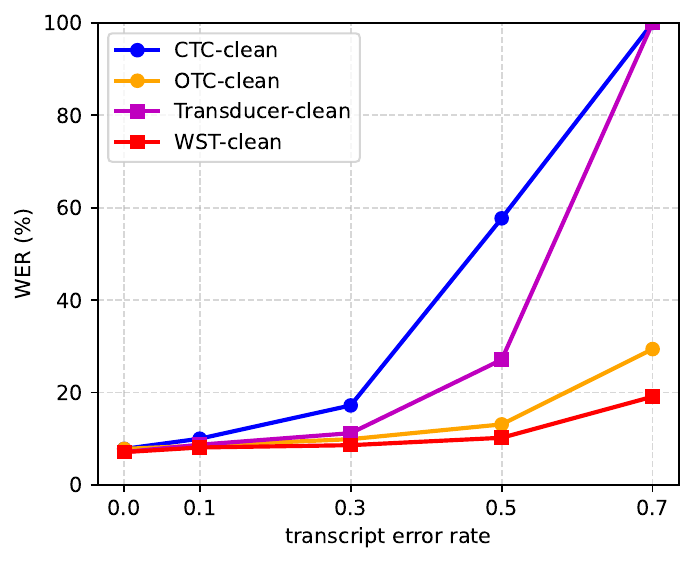}
        \caption{Mixture}
    \end{subfigure}
    \caption{WER ($\%$) (greedy decoding) on LibriSpeech \textit{test-clean} dataset. The model is trained on \texttt{train-clean-100} with synthetic transcript errors: substitution, insertion, deletion, and a mixture of these at error rates: $\{0.0, 0.1, 0.3, 0.5, 0.7\}$. BTC results are compared in (a) substitution and (b) insertion. Results of CTC, BTC, OTC, Transducer, and WST are depicted in \textcolor{blue}{blue}, \textcolor{olive}{green}, and \textcolor{orange}{orange}, \textcolor{purple}{purple}, and \textcolor{red}{red}, respectively.}
    \label{fig:wer_clean_100}
\end{figure}

We begin by conducting simulation experiments on the LibriSpeech dataset to evaluate the effectiveness of the WST approach and to examine its sensitivity to hyperparameter tuning. We conducted four separate tests, each targeting a specific type of transcript error: substitution, insertion, deletion, and a mixture of all three. For each condition, we trained the model using different training criteria—``token bypass only'', ``blank bypass only'', and both combined—and evaluated performance on the LibriSpeech \texttt{test-clean} and \texttt{test-other} set. We evaluate performance using word error rate (WER) obtained through greedy decoding. The results are summarized in Table~\ref{librispeech_wer_clean_wst} and visualized in Figure~\ref{fig:wer_clean_100} for clearer comparison.

\begin{table}[t]
\centering
\begin{tabular}{lllllllllll}
\bottomrule
\multirow{2}{*}{Error}   & \multicolumn{1}{c}{\multirow{2}{*}{Criterion}} & \multicolumn{5}{c}{$p_\text{sub}, p_\text{ins}, p_\text{sub+ins} $}                                                                        \\ \cline{3-7} 
   & \multicolumn{1}{c}{}      & \multicolumn{1}{c}{0.0}      & \multicolumn{1}{c}{0.1} & \multicolumn{1}{c}{0.3} & \multicolumn{1}{c}{0.5} & \multicolumn{1}{c}{0.7} \\ \bottomrule 
\rowcolor{Gainsboro!60}
\multirow{5}{*}{sub}     & CTC             &7.8        &15.1  & 20.8   &47.7  &-             \\
                         & BTC             &7.8               &14.7  & 17.5   &19.8  &-    \\
                         & OTC             &7.8               &8.9  & 11.0   &15.4  &21.5    \\ 
\rowcolor{Gainsboro!60}
                        & Transducer      &7.1               &9.5  & 12.0   &29.4  &-             \\
                         & WST   &7.1                &8.3 &9.0 &10.4 &13.0 \\       \bottomrule
\rowcolor{Gainsboro!60}
\multirow{5}{*}{ins}     & CTC            &7.8          &18.7   &29.8     &72.8      &-             \\
                         & BTC             &7.8         &12.0   &12.1     &12.1   &12.7     \\ 
                         & OTC             &7.8         &7.8   &7.9     &7.9   &8.0     \\ 
\rowcolor{Gainsboro!60}
                         & Transducer      &7.1         &8.6   &9.5     &11.2      &14.6             \\
                         & WST   &7.1                   &7.3  & 7.5    & 7.7  &7.8 \\ \hline
\rowcolor{Gainsboro!60}
\multirow{5}{*}{del}     & CTC             &7.8  &9.8   &17.2     &57.7      &-             \\
                         & BTC             &7.8         &N/A   &N/A     &N/A   &N/A     \\ 
                         & OTC             &7.8         &8.3   &10.3    &17.6   &22.9     \\ 
\rowcolor{Gainsboro!60}
                        & Transducer       &7.1        &9.2   &21.7     &-      &-             \\
                         & WST   &7.1                   &8.3   &10.4    &15.8   &21.6 \\ \hline
\rowcolor{Gainsboro!60}
\multirow{5}{*}{mixed}   & CTC          &7.8                &10  &17.2  &-  & -            \\
                         & BTC            &7.8              &N/A  &N/A  &N/A  &N/A             \\ 
                         & OTC            &7.8              &8.5  &9.9  &13.1  &29.4             \\ 
\rowcolor{Gainsboro!60}
                         & Transducer    &7.1               &8.7  &11.2  &27.1  & -            \\
                         & WST   &7.1                       &8.1  &8.6  &10.2 &19.1\\ \bottomrule
\end{tabular}
\caption{WER ($\%$) (greedy decoding) on LibriSpeech \textit{test-clean} dataset. We compared CTC, Transducer (highlighted in grey) BTC, OTC, and WST in three scenarios: substitution-only, insertion-only, deletion-only and mixed cases. We measure using greedy search. ``-'' indicates that the model does not converge. "N/A" indicates not applicable, since BTC cannot handle deletion errors.}
\label{librispeech_wer_clean_wst}
\end{table}

\begin{table}[t]
\centering
\begin{tabular}{llllllllll}
\bottomrule
\multirow{2}{*}{Error}   & \multicolumn{1}{c}{\multirow{2}{*}{Criterion}} & \multicolumn{5}{c}{$p_\text{sub}, p_\text{ins}, p_\text{sub+ins} $}                                                                        \\ \cline{3-7} 
   & \multicolumn{1}{c}{}    & \multicolumn{1}{c}{0.0}       & \multicolumn{1}{c}{0.1} & \multicolumn{1}{c}{0.3} & \multicolumn{1}{c}{0.5} & \multicolumn{1}{c}{0.7} \\   \\ \bottomrule 
\rowcolor{Gainsboro!60}
\multirow{5}{*}{sub}     & CTC                 &19.4           &29.2  &36.5    &60.5  &-             \\
                         & BTC              &19.4              &29.0  & 33.3   &36.5  &-    \\
                         & OTC               &19.4             &21.2  &24.7   & 32.3  & 39.1   \\ 
\rowcolor{Gainsboro!60}
                         & Transducer       &17.8              &22.7  &29.3    &40.5  &-             \\
                         & WST   &17.8                         &20.1  &21.8  &23.6  &26.9 \\ \bottomrule
\rowcolor{Gainsboro!60}
\multirow{5}{*}{ins}     & CTC         &19.4                    &31.0   &44.3     &-      &-             \\
                         & BTC           &19.4                  &24.0   &24.1     &24.4   &24.6     \\ 
                         & OTC          &19.8                  &19.4  & 19.5    & 19.6  & 21.4    \\ 
\rowcolor{Gainsboro!60}
                         & Transducer     &17.8                 &20.6   &22.1     &25.1      &28.5             \\
                         & WST   &17.8                          &18.1  &18.2 &18.9 &19.3\\ \hline
\rowcolor{Gainsboro!60}
\multirow{5}{*}{del}     & CTC        &19.4                     &21.5   & 29.4     & -     &-             \\
                         & BTC        &19.4                     &N/A   &N/A     &N/A   &N/A     \\ 
                         & OTC          &19.4                   &20.7   &23.7    &33.2   &41.8     \\ 
\rowcolor{Gainsboro!60}
                         & Transducer     &17.8                 &20.5   & 31.5     & -     &-             \\
                         & WST   &17.8                          &20.3 &24.0 &28.7  &39.5 \\ \hline
\rowcolor{Gainsboro!60}
\multirow{5}{*}{mixed}   & CTC     &19.4                     &21.5  &28.0  &-  & -            \\
                         & BTC           &19.4               &N/A &N/A  &N/A  &N/A             \\ 
                         & OTC           &19.4               &20.4  &22.3  &26.4  & 44.6            \\ 
\rowcolor{Gainsboro!60}
                         & Transducer   &17.8                &21.0  &23.7  &37.1  & -            \\
                         & WST  &17.8                        &19.2  &20.2  &24.2 & 30.4\\ \bottomrule
\end{tabular}
\caption{WER ($\%$) (greedy decoding) on LibriSpeech \textit{test-other} dataset. We compared CTC, Transducer (highlighted in grey) BTC, OTC, and WST in three scenarios: substitution-only, insertion-only, deletion-only and mixed cases. We measure using greedy search. ``-'' indicates that the model does not converge. "N/A" indicates not applicable, since BTC cannot handle deletion errors..}
\label{librispeech_wer_other_wst}
\end{table}

\begin{figure}[t]
\centering
    \begin{subfigure}[b]{0.49\linewidth}
        \centering
        \includegraphics[width=\linewidth]{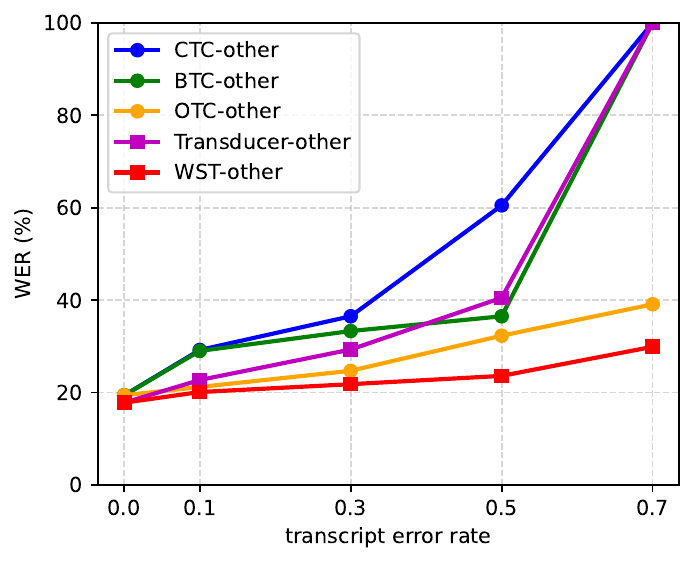}
        \caption{Substitution}
        \vspace{5mm}
    \end{subfigure}
    \begin{subfigure}[b]{0.49\linewidth}
        \centering
        \includegraphics[width=\linewidth]{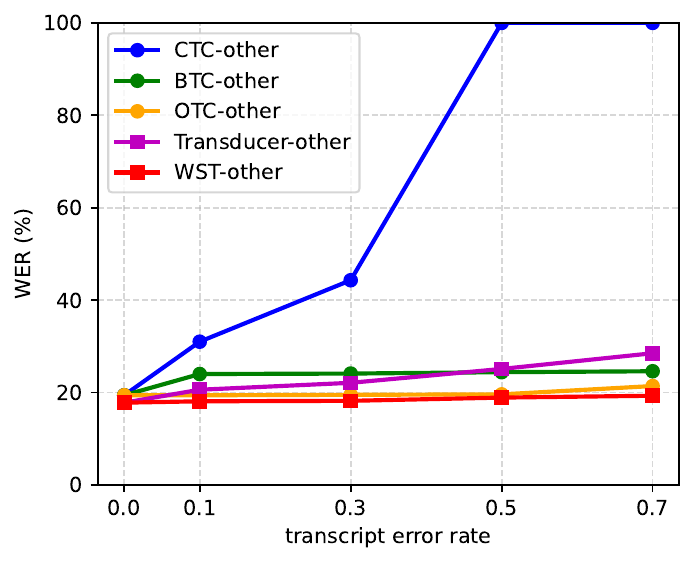}
        \caption{Insertion}
        \vspace{5mm}
    \end{subfigure}\hfill
    \begin{subfigure}[b]{0.49\linewidth}
        \centering
        \includegraphics[width=\linewidth]{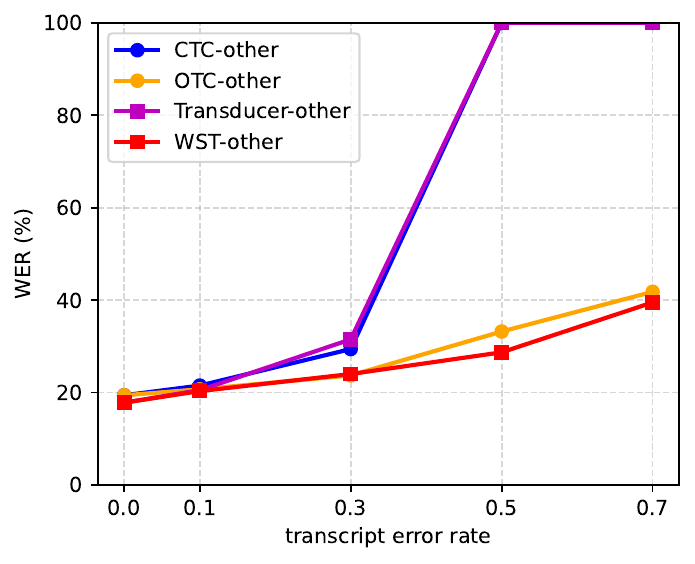}
        \caption{Deletion}
    \end{subfigure}
    \begin{subfigure}[b]{0.49\linewidth}
        \centering
        \includegraphics[width=\linewidth]{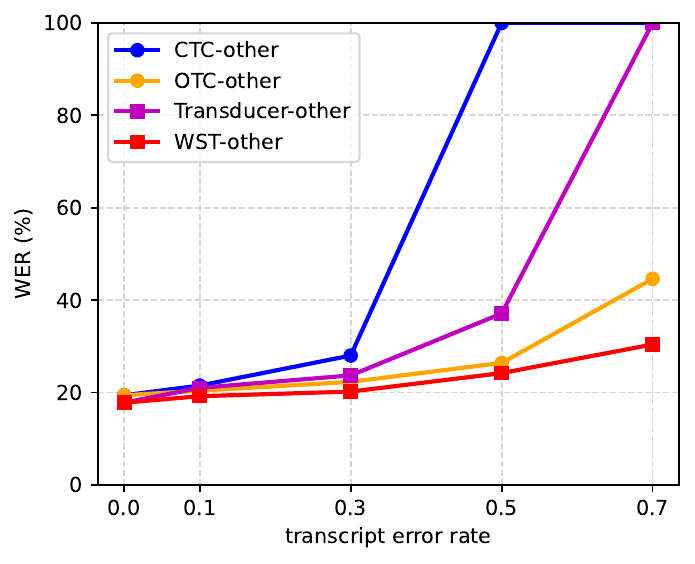}
        \caption{Mixture}
    \end{subfigure}
    \caption{WER ($\%$) (greedy decoding) on LibriSpeech \textit{test-clean} dataset. The model is trained on \texttt{train-clean-100} with synthetic transcript errors: substitution, insertion, deletion, and a mixture of these at error rates: $\{0.0, 0.1, 0.3, 0.5, 0.7\}$. BTC results are compared in (a) substitution and (b) insertion. Results of CTC, BTC, OTC, Transducer, and WST are depicted in \textcolor{blue}{blue}, \textcolor{olive}{green}, and \textcolor{orange}{orange}, \textcolor{purple}{purple} and \textcolor{red}{red}, respectively.}
    \label{fig:wer_other_100}
\end{figure}

\begin{enumerate}
\item \textbf{WST vs. CTC under perfect supervision ($p=0.0$)}: the Transducer and WST both achieve a WER of $7.1\%$ on the \texttt{test-clean} set and $17.8\%$ on the \texttt{test-other} set. This confirms that WST preserves clean-signal performance despite its more flexible training graph. 

\item \textbf{Transducer vs. CTC Across All Error Scenarios}: As in the \texttt{test-clean} results, both CTC and the Transducer experience increasing WERs as noise intensifies. However, the Transducer consistently degrades more slowly, indicating superior robustness. In particular, under the insertion error setting—where $70\%$ of the tokens are incorrectly inserted into the transcripts—CTC fails to converge, whereas the Transducer model is still able to achieve a WER of 14.6 in \texttt{test-clean} and  28.5 in \texttt{test-other}.

\item \textbf{BTC and OTC vs. CTC}: 
Both BTC and OTC consistently outperform CTC across all error types for which they are applicable. However, BTC is unable to handle deletion or mixed noise (indicated by ``N/A'') due to its limited graph structure.

\item \textbf{WST vs. OTC}: 
WST consistently outperforms OTC across all noise conditions. For insertion and deletion errors, WST achieves slightly better performance than OTC across all levels of synthetic noise. In the case of substitution errors, however, the advantage is more substantial, with the performance gap widening as noise severity increases. For example, when $70\%$ of the tokens are incorrect, WST achieves a substantially lower word error rate (WER) of 13.0 on the \texttt{test-clean} set, compared to 21.5 for OTC—representing a relative reduction of nearly $40\%$. A similar improvement is observed on the \texttt{test-other} set, where WST reduces the WER from 39.1 to 26.9.

\item \textbf{WST Achieves the Best Performance Overall}:
Across the entire error types and severity levels, WST yields the lowest WERs in nearly all noisy conditions. It maintains clean-data performance on par with standard Transducer, while offering significantly better robustness as token noise increases. These results establish WST as the most reliable approach for ASR training with noisy transcripts.
\end{enumerate}

\subsection{IH-10k}
We evaluate the standard Transducer and WST, both trained on the IH-10k dataset, using an internal benchmark. Note that, despite considerable cleaning efforts, the IH-10k dataset remains somewhat noisy, with approximately $10\%$ of the transcripts still containing errors.

\begin{table}[t]
\begin{tabular}{llll}
\toprule
Subset             & Transducer     & WST            & \begin{tabular}[c]{@{}l@{}}Relative Improvement \end{tabular} \\ \toprule
\textbf{en-Accent-1}     & \textbf{19.78} & \textbf{18.93} & 4.32\hspace{1mm} \tikz[baseline]{\fill[green!70] (0,0) rectangle (\dimexpr 0.13\linewidth, 0.8em);}                           \\

\textbf{en-Accent-2}     & \textbf{23.98} & \textbf{22.75} & 5.12\hspace{1mm} \tikz[baseline]{\fill[green!70] (0,0) rectangle (\dimexpr 0.16\linewidth, 0.8em);}                           \\ 
\textbf{en-Accent-3}     & \textbf{16.35} & \textbf{14.76} & 9.71\hspace{1mm} \tikz[baseline]{\fill[green!70] (0,0) rectangle (\dimexpr 0.29\linewidth, 0.8em);}                           \\ 
%{\color[HTML]{656565} dictation}          & 11.64          & 10.62          & 8.72\hspace{1mm} \tikz[baseline]{\fill[green!70] (0,0) rectangle (\dimexpr 0.26\linewidth, 0.8em);}    \\
%{\color[HTML]{656565} call center}        & 28.88          & 25.77          & 10.78\hspace{0.4mm}\tikz[baseline]{\fill[green!70] (0,0) rectangle (\dimexpr 0.32\linewidth, 0.8em);}  \\ \midrule
\textbf{en-Accent-4}     & \textbf{28.78} & \textbf{27.27} & 5.25\hspace{1mm} \tikz[baseline]{\fill[green!70] (0,0) rectangle (\dimexpr 0.16\linewidth, 0.8em);}                           \\ 
%{\color[HTML]{656565} dictation}          & 18.25          & 17.07          & 6.49\hspace{1mm} \tikz[baseline]{\fill[green!70] (0,0) rectangle (\dimexpr 0.20\linewidth, 0.8em);}    \\
%{\color[HTML]{656565} call center}        & 39.50          & 37.66          & 4.67\hspace{1mm} \tikz[baseline]{\fill[green!70] (0,0) rectangle (\dimexpr 0.14\linewidth, 0.8em);}    \\ \midrule
\textbf{en-Accent-5}     & \textbf{24.48} & \textbf{24.50} & -0.10 \tikz[baseline]{\fill[red!70] (0,0) rectangle (\dimexpr 0.03\linewidth, 0.8em);}                            \\ 
\textbf{en-Accent-6} & \textbf{22.79}          & \textbf{21.87}          & 4.03\hspace{1mm} \tikz[baseline]{\fill[green!70] (0,0) rectangle (\dimexpr 0.12\linewidth, 0.8em);}                           \\ \hline
%{\color[HTML]{656565} conversation}       & 20.54          & 19.28          & 6.14\hspace{1mm} \tikz[baseline]{\fill[green!70] (0,0) rectangle (\dimexpr 0.18\linewidth, 0.8em);}    \\
%{\color[HTML]{656565} dictation}          & 22.93          & 22.04          & 3.91\hspace{1mm} \tikz[baseline]{\fill[green!70] (0,0) rectangle (\dimexpr 0.12\linewidth, 0.8em);}    \\ \midrule
\textbf{Total}     & \textbf{22.65} & \textbf{21.71} & 4.15\hspace{1mm} \tikz[baseline]{\fill[green!70] (0,0) rectangle (\dimexpr 0.12\linewidth, 0.8em);}                           \\ \bottomrule
\end{tabular}
\caption{WER (\%) on internal benchmark: We compare the performance of the Transducer and WST models across various scenarios. \textcolor{green}{Green} highlights the relative improvement of WST compared to the Transducer, while \textcolor{red}{Red} indicates cases where WST shows a relative performance loss compared to the Transducer.}
\label{table:ih-10k}
\end{table}

The internal benchmark is for evaluating ASR systems under real-world deployment conditions. It spans multiple regional English varieties and includes diverse acoustic environments, making it a valuable testbed for robustness evaluation.

Table~\ref{table:ih-10k} presents word error rate (WER) results for both the standard Transducer and the proposed WST across several subsets: \textbf{en-Accent-1}, \textbf{en-Accent-3}, \textbf{en-Accent-2}, \textbf{en-Accent-4}, \textbf{en-Accent-5}, and \textbf{en-Accent-6} (a mixed group of speakers with non-native or regionally unclassified English accents).

WST outperforms the standard Transducer in five of the six subsets, with relative improvements ranging from modest ($4.03\%$) to substantial ($9.71\%$). A breakdown by region follows:

\begin{itemize}
    \item \textbf{en-Accent-1:} WST reduces WER from $19.78\%$ to $18.93\%$, a relative improvement of $4.32\%$. 
    \item \textbf{en-Accent-2:} WST achieves a $5.12\%$ relative improvement, lowering WER from $23.98\%$ to $22.75\%$.
    \item \textbf{en-Accent-3:} The largest gain is observed on this subset, where WER drops from $16.35\%$ to $14.76\%$, corresponding to a $9.71\%$ relative improvement. 
    \item \textbf{en-Accent-4:} For this subset, WST achieves a $5.25\%$ improvement, reducing WER from $28.78\%$ to $27.27\%$.
    \item \textbf{en-Accent-5:} This is the only subset where WST slightly underperforms, with WER increasing from $24.48\%$ to $24.50\%$ (a relative drop of $0.10\%$). The difference is marginal.
    \item \textbf{en-Accen-6:} This category includes non-native and regionally ambiguous English. WST improves WER from $22.79\%$ to $21.87\%$, yielding a $4.03\%$ relative gain.
\end{itemize}

\textbf{Overall}, WST reduces the total WER across the full benchmark from $22.65\%$ to $21.71\%$, representing a \textbf{$4.15\%$ relative improvement}. This consistent performance gain across varied accents and domains reinforces the practical value of WST for large-scale deployment, especially when handling imperfect supervision and regional speech variation.

\section{Relation to Concurrent Work}
This work was developed independently, though its similarity to a concurrent technical report\footnote{\url{https://arxiv.org/pdf/2504.06963}} may be partially attributed to prior informal discussions between the author and ourselves regarding the general idea. We became aware of the report only after completing this work. While both studies adopt similar strategies for transducer-based training under noisy supervision using a WFST framework, they differ in several aspects:
\begin{itemize}
    \item The technical report explores a setup in which a constant weight/penalty is used for the “skip-frame” arc, and the sum of the probabilities of all labels excluding the blank and ground-truth labels is combined with a decaying penalty for “bypass” arcs. In this paper, we define the weights as the logarithm of the average probability of all non-blank tokens, plus a tunable penalty for both types of arcs, consistent with OTC. We also remove the decay component, as we find that it introduces an additional hyperparameter with minimal effect on the results.
    
    \item The report simulates settings with 20\% and 50\% transcription errors, whereas this paper evaluates a broader range of noise levels: 10\%, 30\%, 50\%, and 70\%.
    
    \item The report presents results only on simulated transcription errors using the LibriSpeech dataset, while this paper includes additional empirical results on a larger (10,000-hour) industrial speech corpus containing naturally occurring and uncurated transcription errors.
\end{itemize}

\section{Conlusions}

In conclusion, the Weakly Supervised Transducer (WST) demonstrates robust and consistent improvements over BTC, OTC, and traditional Transducer models across diverse datasets and scenarios. 
By effectively handling weak supervision and noisy transcripts, WST achieves significant reductions in WER, particularly in challenging environments such as call centers and multi-accent conversational speech. 
While minor performance degradations are observed in specific contexts, the overall gains highlight WST's adaptability and effectiveness in real-world, large-scale ASR applications. 
These results underscore the potential of WST to advance the development of more resilient and accurate speech recognition systems.

\newpage
\bibliographystyle{IEEEtran}
\bibliography{mybib}

\end{document}